\pdfoutput=1

\documentclass[11pt]{article}

\usepackage[]{acl}

\usepackage{times}
\usepackage{latexsym}

\usepackage[T1]{fontenc}

\usepackage[utf8]{inputenc}

\usepackage{algorithm}
\usepackage{algorithmic}

\usepackage{makecell}

\usepackage{hyperref}
\usepackage{color}
\definecolor{span_pink}{RGB}{255,122,179}
\definecolor{entity_blue}{RGB}{148,169,216}
\usepackage{newfloat}
\usepackage{listings}
\floatstyle{ruled}
\newfloat{listing}{tb}{lst}{}
\floatname{listing}{Listing}

\usepackage{microtype}
\usepackage{booktabs} 
\usepackage{url}
\usepackage{float}
\usepackage{graphicx}
\usepackage{multirow}
\usepackage{multicol}
\usepackage{balance}
\usepackage{amsfonts}
\usepackage{amsmath}
\usepackage{xcolor}
\usepackage{diagbox}
\usepackage{subfigure}
\makeatletter
\newcommand{\thickhline}{%
    \noalign {\ifnum 0=`}\fi \hrule height 1pt
    \futurelet \reserved@a \@xhline
}
\newcommand*\circled[1]{\kern-2.5em%
  \put(0,4){\color{white}\circle*{18}}\put(0,4){\circle{10}}%
  \put(-3,0){\color{black}\bfseries#1}~~}
\usepackage{tikz} 

\usepackage{enumitem}

\newcommand{\modelname}{\texttt{HiURE}}
\newcommand{\encoder}{Contextualized Relation Encoder}
\newcommand{\contrastive}{Hierarchical Exemplar Contrastive Learning}
\newcommand{\printfnsymbol}[1]{%
  \textsuperscript{\@fnsymbol{#1}}%
}
\title{HiURE: Hierarchical Exemplar Contrastive Learning\\ for Unsupervised Relation Extraction}

\author{Shuliang Liu{$^1$}\thanks{~~Equal contribution.},~Xuming Hu{$^1$}\footnotemark[1],~Chenwei Zhang{$^2$}\footnotemark[2], ~Shu'ang Li$^1$, ~Lijie Wen$^1$\thanks{~ Corresponding authors.},  ~Philip S. Yu$^{1, 3}$\\
  {$^1$Tsinghua University}\\ {$^2$Amazon}\\
    {$^3$University of Illinois at Chicago}\\
  {\tt \{liusl19,hxm19,lisa18\}@mails.tsinghua.edu.cn}\\
  {\tt cwzhang@amazon.com};~{\tt wenlj@tsinghua.edu.cn};~{\tt psyu@uic.edu} 
}

\begin{document}
\maketitle
\begin{abstract}
Unsupervised relation extraction aims to extract the relationship between entities from natural language sentences without prior information on relational scope or distribution. Existing works either utilize self-supervised schemes to refine relational feature signals by iteratively leveraging adaptive clustering and classification that provoke gradual drift problems, or adopt instance-wise contrastive learning which unreasonably pushes apart those sentence pairs that are semantically similar. To overcome these defects, we propose a novel contrastive learning framework named {\modelname}, which has the capability to derive hierarchical signals from relational feature space using cross hierarchy attention and effectively optimize relation representation of sentences under exemplar-wise contrastive learning. Experimental results on two public datasets demonstrate the advanced effectiveness and robustness of {\modelname} on unsupervised relation extraction when compared with state-of-the-art models. Source code is available here\footnote{\url{https://github.com/THU-BPM/HiURE}}.
\end{abstract}

\section{Introduction}\label{introduction}
Relation Extraction (RE) aims to discover the semantic (binary) relation that holds between two entities from plain text. For instance, ``Kissel$_{head}$ was born in Adrian$_{tail}$ ...", we can extract a relation ${\texttt{/people/person/place\_of\_birth}}$ between the two head-tail entities. The extracted relations could be used in various downstream applications such as information retrieval \cite{corcoglioniti2016knowledge}, question answering \cite{bordes2014question}, and dialog systems \cite{madotto2018mem2seq}.

\begin{figure*}[t!]
    \centering
    \includegraphics[width=0.95\linewidth]{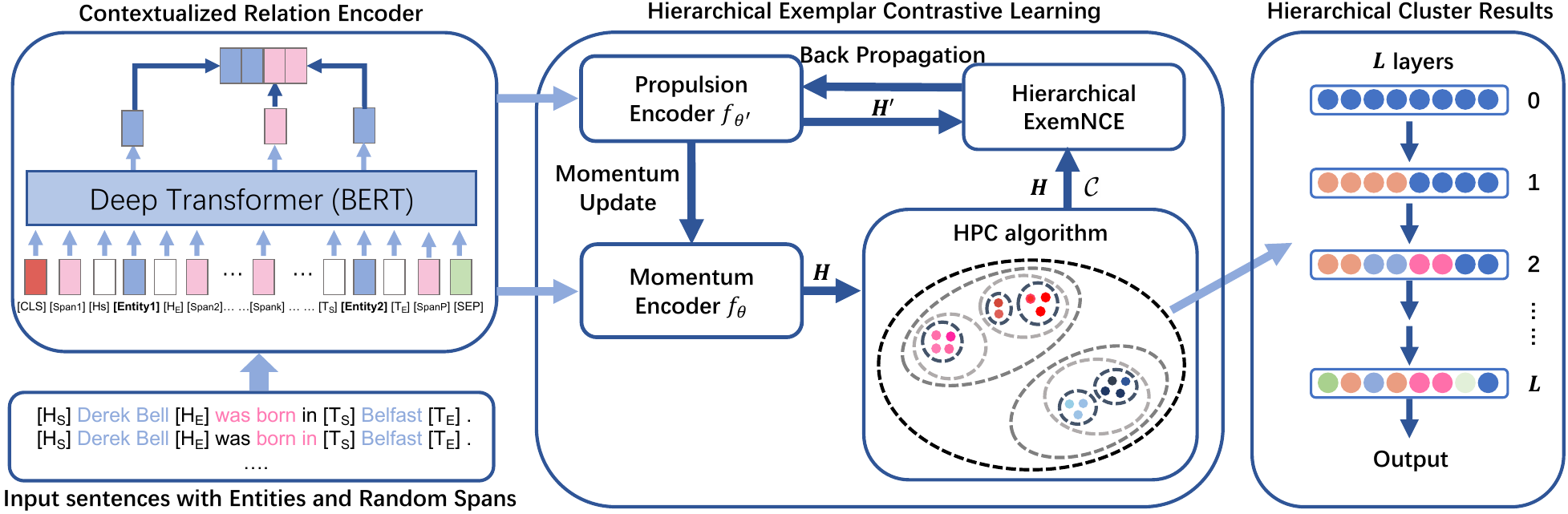}
    \caption{Framework of \modelname. Sentence representations will be augmented through {\textcolor{span_pink}{Random Spans}} with fixed {\textcolor{entity_blue}{Entities}}, then transmitted into Propulsion and Momentum Encoder respectively. The HPC algorithm takes Momentum feature $H$ as input and generates $L$ layers of clustering results together with $L$ exemplar sets $\mathcal{C}$. HiNCE takes $H$ and  $H^{\prime}$ for instance-wise while $H$ and $\mathcal{C}$ for exemplar-wise contrastive learning.}
    
    \label{fig:overview}
\vspace{-0.1in}
\end{figure*}

Existing RE methods can achieve decent results with manually annotated data or human-curated knowledge bases. While in practice, human annotation can be labor-intensive to obtain and hard to scale up to newly created relations. Lots of efforts are devoted to alleviating the impact of human annotations in relation extraction. Unsupervised Relation Extraction (URE) is especially promising since it does not require any prior information on relation scope and distribution. The main challenge in URE is how to cluster semantic information of sentences in the relational feature space. 

\citet{simon2019unsupervised} adopted skewness and dispersion losses to enforce relation classifier to be confident in the relational feature prediction and ensure all relation types can be predicted averagely in a minibatch. But it still requires the exact number of relation types in advance, and the relation classifier could not be improved by obtained clustering results. \citet{hu-etal-2020-selfore} encoded relational feature space in a self-supervised method that bootstraps relational feature signals by leveraging adaptive clustering and classification iteratively. Nonetheless, like other self-training methods, the noisy clustering results will iteratively result in the model deviating from the global minima, which is also known as gradual drift problem \citep{curran2007minimising,zhang2016understanding}.

\citet{peng2020learning} leveraged contrastive learning to obtain a flat metric for sentence similarity in a relational feature space. However, it only considers the relational semantics in the feature space from an instance perspective, which will treat each sentence as an independent data point. As scaling up to a larger corpus with potentially more relations in a contrastive learning framework, it becomes more frequent that sentence pairs sharing similar semantics are undesirably pushed apart in a flat relational feature space. Meanwhile, we observe that many relation types can be organized in a hierarchical structure. For example, the relations \texttt{/people/person/place\_of\_birth} and \texttt{/people/family/country} share the same parent semantic on \texttt{/people}, which means that they belong to the same semantic cluster from a hierarchical perspective. Unfortunately, these two relations will be pushed away from each other in an instance-wise contrastive learning framework.

Therefore, our intuitive approach is to alleviate the dilemma of similar sentences being pushed apart in contrastive learning by leveraging the hierarchical cluster semantic structure of sentences. 
Nevertheless, traditional hierarchical clustering methods all suffer from the gradual drift problem. Thereby, we try to exploit a new approach of hierarchical clustering by combining propagation clustering and attention mechanism.
We first define \textbf{exemplar} as a representative instance for a group of semantically similar sentences in certain clustering results. Exemplars can be in different granularities and organized in a hierarchical structure. 
In order to enforce relational features to be more similar to their corresponding exemplars in all parent granularities than others, we propose {\modelname}, a novel contrastive learning framework for URE which combines both the instance-wise and exemplar-wise learning strategies, to gather more reasonable relation representations and better classification results. 

The proposed {\modelname} model is composed of two modules: {\encoder} and Hierarchical Exemplar Contrastive Learning. As shown in Figure \ref{fig:overview}, the encoder module leverages pre-trained BERT model to obtain two augmented entity-level relational features of each sentence for instance-wise contrastive learning,
while the learning module retrieves hierarchical exemplars in a top-down fashion for exemplar-wise contrastive learning and updates the features of sentences iteratively according to the hierarchy. These updated features could be utilized to optimize the parameters of encoders by a combined loss function noted as Hierarchical ExemNCE (HiNCE) in this work. To summarize, the main contributions of this paper are as follows:
\begin{itemize}
\item We develop a novel hierarchical exemplar contrastive learning framework {\modelname} that incorporates top-down hierarchical propagation clustering for URE.
\vspace{-0.15in}
\item We demonstrate how to leverage the semantic structure of sentences to extract hierarchical relational exemplars which could be used to refine contextualized entity-level relational features via HiNCE.
\vspace{-0.15in}
\item We conduct extensive experiments on two datasets and {\modelname} achieves better performance than the existing state-of-the-art methods. This clearly shows the superior capability of our model for URE by leveraging different types of contrastive learning. Our ablation analysis also shows the impacts of different modules in our framework.
\end{itemize}

\section{Proposed Model}
The proposed model {\modelname} consists of two modules: {\encoder} and {\contrastive}. As illustrated in Figure \ref{fig:overview}, the encoder module takes natural language sentences as input, where named entities are recognized and marked in advance, then employs the pre-trained BERT \cite{devlin2019bert} model to output two contextualized entity-level feature sets $H$ and $H^{\prime}$ for each sentence based on Random Span. The learning module takes these features as input, and aims to retrieve exemplars that represent a group of semantically similar sentences in different granularities, denoted as $\mathcal{C}$. We leverage these exemplars to iteratively update relational features of sentences in a hierarchy and construct an exemplar-wise contrastive learning loss called Hierarchical ExemNCE which enforces the relational feature of a sentence to be more similar to its corresponding exemplars than others.
\subsection{\encoder}
The {\encoder} aims to obtain two relational features from each sentence based on the context information of two given entity pairs for instance-wise contrastive learning. In this work, we assume named entities in the sentence have been recognized in advance.

For a sentence ${x = [w_{1}, ..,w_{T}]}$ with $T$ words where each $w_i$ represents a word and two entities Head and Tail are mentioned, we follow the labeling schema adopted in \citet{soares2019matching} and argument ${x}$ with four reserved tokens to mark the beginning and the end of each entity. We introduce ${[\operatorname{H}_{\text{S}}]}$, ${\operatorname{[H}_{\text{E}}]}$, ${[\operatorname{T}_{\text{S}}]}$, ${[\operatorname{T}_{\text{E}}]}$ to represent the start or end position of head or tail entities respectively and inject them to $x$:
\begin{equation}
\begin{split}
{x^{\prime}}=&\big[w_{1},...,[\operatorname{H}_{\text{S}}],...,w_{i},...,[\operatorname{H}_{\text{E}}],...,w_{Span1},...,\\
&\quad w_{SpanP},...,{[\operatorname{T}_{\text{S}}],...,w_{j},...,[\operatorname{T}_{\text{E}}],...,w_{T}\big]}
\end{split}
\end{equation}
where $x^{\prime}$ will be the input token sequence for the encoder and $Span$ subscript indicates the Random Span words. Considering the relational features between entity pairs are normally embraced in the context, we use pre-trained BERT \citep{devlin2019bert} model to effectively encode every tokens in the sentence along with their contextual information, and get the token embedding $\mathbf{b}_i=f_{\text{BERT}}(w_i)$, where $i\in[1,T]$ including the special tokens in $x^{\prime}$ and $\mathbf{b}_i \in \mathbb{R}^{\cdot{b_{R}}}$, where $b_{R}$ represents the dimension of the token embedding.

We utilize the outputs $\mathbf{b}_i$ corresponding to ${[\operatorname{H}_{\text{S}}]}$ and ${[\operatorname{T}_{\text{S}}]}$ as the contextualized entity-level features instead of using sentence-level marker $[CLS]$ to get embedding for target entity pair. For contrastive learning purposes, we randomly select $P$ words as \textbf{Random Span} from all the context words in the whole sentence except for those entity words 
and special tokens to augment the entity-level features as $\mathbf{b}_{Span}$, where multiple different Random Span selections lead to different semantically invariant embedding of the same sentence. For every selection, we concatenate the  embedding of the two entity and $P$ Random Span words together to derive a fixed-length relational feature $\mathbf{h}\in\mathbb{R}^{\left(2+P\right)\cdot{b_{R}}}$:
\begin{equation}\label{h}
\mathbf{h} = [\mathbf{b}_{[\operatorname{H}_{\text{S}}]}, \mathbf{b}_{[\operatorname{T}_{\text{S}}]}, \mathbf{b}_{Span1},..., \mathbf{b}_{SpanP}]
\end{equation}
where $\mathbf{h}$ is the output of the {\encoder} which can be denoted as $f_{\theta}(x,\operatorname{Head},\operatorname{Tail},Span)$.  The Random Span strategy can get sentence-level enhanced relational features to construct positive samples directly and effectively, and its simplicity highlights the role of subsequent modules.
\subsection{\contrastive} 
In order to adaptively generate more positive samples other than sentences themselves to introduce more similarity information in contrastive learning, we design hierarchical propagation clustering to obtain multi-level cluster exemplars as positive samples of corresponding instances.

We assume the relation hierarchies are tree-structured and define hierarchical exemplars as representative relational features for a group of semantically similar sentences with different granularities. The exemplar-wise contrastive learning encourages relational features to be more similar to their corresponding exemplars than other exemplars. 

The process is completed through \textit{Hierarchical Propagation Clustering} (HPC) to generate cluster results of different granularities and \textit{Hierarchical Exemplar Contrastive Loss} (HiNCE) to optimize the encoder. The main procedure of HPC consists of Propagation Clustering and \textit{Cross Hierarchy Attention} (CHA), as is elaborated in Algorithm \ref{alg:pc_algorithm}, which will be explained in detail below.

\noindent\textbf{Propagation Clustering}\\
\begin{algorithm}[tb!]
\caption{Hierarchical Propagation Clustering}
\label{alg:pc_algorithm}
\textbf{Input}: Encoder outputs ${H = \{\mathbf{h}_{1},\mathbf{h}_{2},...,\mathbf{h}_{n}\}}$, \\
\hspace*{1cm} Hierarchical cluster layers $L$  \\
\mbox{ \textbf{Output}: Hierarchical clusters results ${\mathcal{C}}$}
\vspace{-1.5em}
\begin{algorithmic}[1] 
\STATE $H^1 \leftarrow H ,\  \mathcal{C} \leftarrow [\ ]$
\label{line:initialize_vars}
\STATE Initialize $\{s_{ij}|i,j \in [1,n]\}$ by Eq. \ref{similarity_cal}
\label{line:initialize_s}
\STATE ${\forall i \neq j}:\ p_{\top }=\min(s_{i j}), \  p_{\bot }=\operatorname{median}(s_{i j})$
\label{line:p_top_and_bot}
\STATE $p s=\left\{p_{l} \mid p_{l}=p_{\top }+\frac{p_{\bot }-p_{\top }}{L-1}\cdot(l-1), l\in{[1,L]} \right\}$
\label{line:generate_prefs}
\vspace{-1.0em}
\FOR{$l$ in $[1,L]$} 
\label{line:hpc_start}
\STATE Update $\{s_{ij}\}$ according to $H^l$ by Eq. \ref{similarity_cal}
\label{line:update_s}
\STATE Set diagonal to preference $s_{ii}=p_l$
\label{line:set_diagonal}
\FORALL{iterations}
\label{line:iterations_start}
\STATE Update $\{r_{ij}\}$ and $\{a_{ij}\}$ by Eq. \ref{r_cal} and \ref{a_cal}
\label{line:update_r_and_a}
\STATE $\hat{\mathbf{c}}\! =\! \left(\hat{c}_{1}, \ldots, \hat{c}_{n}\right), \hat{c}_{i}\! =\! \operatorname{argmax}_{j}(a_{ij}+r_{ij})$
\vspace{-1.0em}
\label{line:cal_exemplar}
\STATE\mbox{\spaceskip=0.2em\relax Exemplar set $E^l\! =\! \{\mathbf{e}^l_{\hat{c}_{i}}|\mathbf{e}^l_{\hat{c}_{i}}\! =\! \mathbf{h}^l_{\hat{c}_{i}},\hat{c}_{i}\in \hat{\mathbf{c}}\}$}
\vspace{-1.0em}
\label{line:exem_set}
\IF{Changes of $E^l$ have converged}
\label{line:converge_condition}
\STATE break
\ENDIF
\label{line:converge_condition_end}
\ENDFOR
\label{line:iterations_end}
\STATE $\mathcal{C}$.add($E^l$)
\label{line:add_exem_set}
\STATE $H^{l+1} \leftarrow (H^l,E^l)$ by Eq. \ref{h_update}
\label{line:update_h}
\ENDFOR
\label{line:hpc_end}
\STATE \textbf{return} $\mathcal{C}$
\label{line:return_c}
\end{algorithmic}
\end{algorithm}
\noindent We use propagation clustering to obtain hierarchical exemplars in an iterative, top-down fashion. Traditional clustering methods such as ${k}$-means cluster data points into specific cluster numbers, however, these methods could not utilize hierarchical information in the dataset and require the specific cluster number in advance. Propagation clustering possesses the following advantages: 1) It considers all feature points as potential exemplars and uses their mutual similarity to extract potential tree-structured clusters. 2) It neither requires the actual number of target relations in advance nor the distribution of relations. 3) It will not be affected by the quality of the initial point selection.

In practice, propagation clustering exchanges real-valued messages between points until a high-quality set of exemplars and corresponding clusters are generated. Inspired by \citet{frey2007clustering}, we adopt similarity $s_{ij}$ to measure the distance between points $i$ and $j$, responsibility $r_{ij}$ to indicate the appropriateness for $j$ to serve as the exemplar for $i$ and availability $a_{ij}$ to represent the suitability for $i$ to choose $j$ as its exemplar:
\begin{equation}\label{similarity_cal}
s_{ij}=-\left \| \mathbf{h}_{i}- \mathbf{h}_{j}\right \| ^{2}
\end{equation}
\vspace*{-1.5em}
\begin{equation}\label{r_cal}
r_{ij}=s_{ij}-\max _{j^{\prime} \neq j}\left(s_{ij^{\prime}}+a_{ij^{\prime}}\right) \\
\end{equation}
\vspace*{-1.5em}
\begin{equation}\label{a_cal}
a_{ij}=\left\{\begin{array}{c}\sum_{i^{\prime} \neq i} \max \left(0, r_{i^{\prime} j}\right),  j=i \\ 
\min \left[0, r_{jj}+\underset{i^{\prime}\notin \left \{ i,j \right \}}{\sum} \max \left(0, r_{i^{\prime} j}\right)\right],
 j \neq i\end{array}\right.
\end{equation}
where $r_{ij}$ and $a_{ij}$ will be updated through the propagation iterations until convergence (Lines~\ref{line:iterations_start}-\ref{line:iterations_end}) and a set of cluster centers , which is called exemplar, will be chosen as $E$ (Line~\ref{line:exem_set}). Then we wish to find a set of $L$ consecutive layers of clustering,  where the points to be clustered in layer $l$ are closer to the corresponding exemplar of layer $l - 1$. We perform propagation clustering $L$ times (Lines \ref{line:hpc_start}-\ref{line:hpc_end}) with different preferences (Lines \ref{line:initialize_s}-\ref{line:generate_prefs}) to generate $L$ different layers of clustering result, where a larger preference leads to more numbers of clusters \cite{moiane2018evaluation}. The Hyperparameter Analysis part provides a detailed explanation about how to select $L$ and the reason for building the preference sequence $ps$ according to the formula in Line \ref{line:generate_prefs}.

\noindent\textbf{Cross Hierarchy Attention}\\
The traditional hierarchical clustering method either merge fine-grained clusters into coarse-grained one or split coarse cluster into fine-grained ones, which will both cause the problem of error accumulation. Preference sequence leads to different cluster results in a hierarchical way but lost the interaction information between adjacent levelsin propagation clustering. Based on this intuition, we introduce CHA mechanism to leverage signals from coarse-grained exemplar to fine-grained clusters.
\begin{figure}[bt!]
    \centering
    \includegraphics[width=0.9\linewidth]{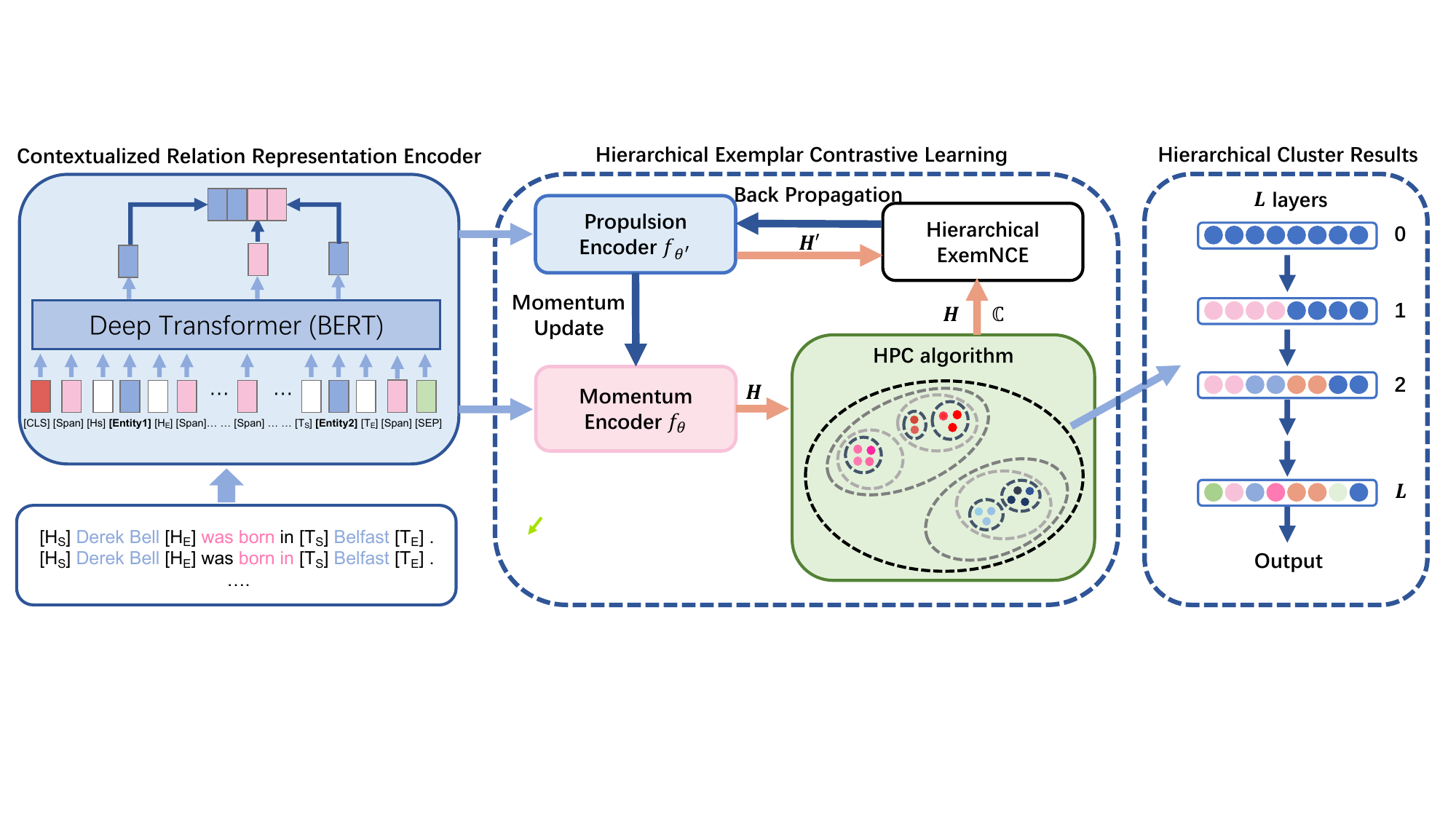}
    \caption{Overview of cross hierarchy attention. The first part shows original data. The second part divides data points into two clusters and utilizes attention to update every points which contribute to the next level of clustering.  The dotted line indicates negative sample pair while solid line with positive in contrastive learning.}
    \label{fig:attention}
\vspace{-0.2in}
\end{figure}

Formally, we derive a CHA matrix $A^l$ at layer $l$ where the element at $(j,k)$ is obtained by a scaled softmax:
\begin{align}
    \alpha^{l}_{jk}=\frac{exp(\lambda \mathbf{e}^{l}_j\cdot \mathbf{e}^{l}_k)}{\sum_{k^{\prime}}exp(\lambda \mathbf{e}^{l}_j\cdot \mathbf{e}^{l}_{k^{\prime}})}
\end{align}
where $\lambda$ is a trainable scalar variable, not a hyper-parameter \cite{luong2015effective}. The attention weight $\alpha^{l}_{jk}$ reflects the proximity between exemplar $j$ and exemplar $k$ in layer $l$ and measures the influence and interactions to corresponding data points between these exemplars. Typically, exemplars that are visually close to each other would have higher attention weights. Then we derive attended point representation at layer $l + 1$ by taking the attention weighted sum of its corresponding exemplar from other exemplars:
\begin{align}
    \hat{\mathbf{h}}^{l+1}_{i} = \sum_{k}\alpha^l_{jk}\mathbf{e}^l_k
\end{align}
where $\mathbf{e}^l_j$ is the exemplar of $\mathbf{h}^{l}_{i}$. The attended representation aggregates signals from other exemplars weighted by how close they are to exemplar $\mathbf{e}^l_j$ and transfer the signals from layer $l$ to $l + 1$. They reflect how likely a neighboring cluster is relevant or the point will get close to it. Then we combine the attended representation with the original one to obtain the CHA based embedding $\mathbf{h}^{l+1}_i$, defined as:
\begin{equation}\label{h_update}
    \mathbf{h}^{l+1}_i = \mathbf{h}^{l}_i + \lambda_{att} \hat{\mathbf{h}}^{l+1}_{i}
\end{equation}
where $\lambda_{att}$ is not a hyper-parameter, but a weighting variable to be automatically trained. As illustrated in Figure \ref{fig:attention}, the CHA mechanism helps data points to get closer with corresponding exemplars in previous layer and thus perform better in the current layer.

\noindent\textbf{Hierarchical Exemplar Contrastive Loss}\\
Given a training set ${X = \{x_{1},x_{2},...,x_{n}\}}$ of ${n}$ sentences, {\encoder} can obtain two augmented relational features for each input sentences by randomly sampling spans twice for the same entity pair. We do this for all sentences and obtain ${H = \{\mathbf{h}_{1},\mathbf{h}_{2},...,\mathbf{h}_{n}\}}$ and ${H^{\prime} = \{\mathbf{h}_{1}^{\prime},\mathbf{h}_{2}^{\prime},...,\mathbf{h}_{n}^{\prime}\}}$. Traditional instance-wise contrastive learning treats two features as a negative pair as long as they are from different instances regardless of their semantic similarity. It updates encoder by optimizing InfoNCE \cite{oord2018representation,peng2020learning}:
\begin{equation}\label{InfoNCE}
    \mathcal{L}_{\rm InfoNCE}=\sum_{i=1}^{n}-\operatorname{log}\frac{\operatorname{exp}(\mathbf{h}_{i}\cdot\mathbf{h}_{i}^{\prime}/\tau )}{\sum_{j=1}^{J}\operatorname{exp}(\mathbf{h}_{i}\cdot\mathbf{h}_{j}^{\prime}/\tau ) }
\end{equation}
where $\mathbf{h}_{i}$ and $\mathbf{h}_{i}^{\prime}$ are positive samples for instance $i$, while $\mathbf{h}_{j}^{\prime}$ includes one positive sample and $J - 1$ negative samples for other sentences, and $\tau$ is a temperature hyper-parameter \cite{wu2018unsupervised}. 

Compared with the traditional instance-wise contrastive learning which unreasonably pushes apart many negative pairs that possess similar semantics, we employ the inherent hierarchical structure in relations.
As illustrated in Figure \ref{fig:overview}, we perform the HPC algorithm iteratively at each epoch to utilize hierarchical relational features.
Note that the relational feature $\mathbf{h}_i$ will be updated in each batch while training, but the exemplars will not be retrieved until the epoch is finished. To maintain the invariance of exemplars and avoid representation shift problems with the relational features in an epoch, we need to smoothly update the parameters of the encoder to ensure a fairly stable relational feature space. In practice, we construct two encoders: Momentum Encoder $f_{\theta}$ and Propulsion Encoder $f_{{\theta}^{\prime}}$, both of which is a instance of the \encoder. ${\theta}^{\prime}$ is updated by contrastive learning loss and ${\theta}$ is a moving average of the updated ${\theta}^{\prime}$ to ensure a smoothly update of relational features \cite{he2020momentum}.
We leverage HPC on the momentum features $\mathbf{h}_{i}=f_{\theta}(x_{i})$ to obtain $\mathcal{C}$ (Line \ref{line:return_c}), which contains $L$ layers of cluster results with $c_l$ exemplars respectively, where $c_l$ is the number of clusters at layer $l$. 
In order to enforce the relational features more similar to their corresponding exemplars compared to other exemplars \cite{caron2020unsupervised,li2020prototypical}, we define exemplar-wise contrastive learning as ExemNCE:
\begin{equation}
    \mathcal{L}_{\rm ExemNCE}\! =\! - \sum_{i=1}^n \frac{1}{L} \sum_{l=1}^L \operatorname{log}\frac{\operatorname{exp}(\mathbf{h}_i\cdot \mathbf{e}^l_j/\tau)}{\sum_{q=1}^{c_l} \operatorname{exp}(\mathbf{h}_i\cdot \mathbf{e}_q^l/\tau)}
\end{equation}
where $j \in [1,c_l]$ and $\mathbf{e}^l_j$ is the corresponding exemplar of instance $i$ at layer $l$. As we have explicitly constrained $\mathbf{h}_{i}$ and $\mathbf{e}^l_j$ into approximate feature space, so the temperature parameter $\tau$ can be shared here. The difference between InfoNCE and ExemNCE is described in the second part of Figure \ref{fig:attention}, where the solid line represents positive while the dashed line represents negative.

Furthermore, we add InfoNCE loss to retain the local smoothness which could help propagation clustering. Overall, our objective named Hierarchical ExemNCE is defined as:
\begin{equation}
\begin{aligned}
\mathcal{L}_{\rm HiNCE}&=\mathcal{L}_{\rm InfoNCE}+\mathcal{L}_{\rm ExemNCE} 
\end{aligned}
\end{equation}

After we update Propulsion Encoder $f_{{\theta}^{\prime}}$ with HiNCE, the Momentum Encoder $f_{\theta}$ can be propulsed by:
\begin{equation}\label{momentum update}
    {\theta}\leftarrow m\cdot{\theta} + (1-m)\cdot{\theta}^{\prime}
\end{equation}
where $m\in \left[ 0,1 \right)$ is a momentum coefficient. The momentum update in Eq. \ref{momentum update} makes ${\theta}$ evolve more smoothly than $\theta^{\prime}$ especially when $m$ is closer to $1$.

\section{Experiments}
We conduct extensive experiments on real-world datasets to prove the effectiveness of our model for Unsupervised Relation Extraction tasks and give a detailed analysis of each module to show the advantages of {\modelname}. Implementation details and evaluation metrics are illustrated in Appendix \ref{impl_details} and \ref{evaluation_metrics} respectively.

\begin{table*}[thbp!]
\centering
  \resizebox{.95\linewidth}{!}{%
\begin{tabular}{clccccccc}
\thickhline
\multirow{2}{*}{\textbf{Dataset}} & \multirow{2}{*}{\textbf{Model}}       & \multicolumn{3}{c}{\textbf{$\text{B}^{3}$}}                       & \multicolumn{3}{c}{\textbf{V-measure}}                                                           & \multirow{2}{*}{\textbf{ARI}}               \\ \cmidrule(lr){3-5}\cmidrule(lr){6-8}
                        &&F1            & Prec.         & Rec.          & F1            & Hom.          & Comp.         &                                    \\ \hline
\multirow{11}{*}{NYT+FB}  & {rel-LDA\cite{yao2011structured}}     & 29.1\small±2.5          & 24.8\small±3.2          & 35.2\small±2.1          & \multicolumn{1}{c}{30.0\small±2.3}          &     \multicolumn{1}{c}{26.1\small±3.3}          & 35.1\small±3.5          & \multicolumn{1}{c}{13.3\small±2.7}          \\ 
                         & March\cite{marcheggiani2016discrete}            & 35.2\small±3.5          & 23.8\small±3.2         & 67.1\small±4.1 & 27.0\small±3.0         & 18.6 \small±1.8         & 49.6\small±3.1         & \multicolumn{1}{c}{18.7\small±2.6}          \\ 
                         & UIE-PCNN\cite{simon2019unsupervised}              & 37.5\small±2.9          & 31.1\small±3.0         & 47.4\small±2.8        & 38.7\small±3.2         & 32.6\small±3.3          & 47.8\small±2.9          & \multicolumn{1}{c}{27.6\small±2.5}          \\ 
                         & UIE-BERT\cite{simon2019unsupervised} & 38.7\small±2.8         & 32.2\small±2.4         & 48.5\small±2.9         & 37.8\small±2.1         & 32.3\small±2.9          & 45.7\small±3.1         & \multicolumn{1}{c}{29.4\small±2.3}          \\ 
                         & SelfORE\cite{hu-etal-2020-selfore} & 41.4\small±1.9         & 38.5\small±2.2          & 44.7\small±1.8         & 40.4\small±1.7          & 37.8\small±2.4        & 43.3\small±1.9          & \multicolumn{1}{c}{35.0\small±2.0}   \\ 
                         & EType\cite{tran-etal-2020-revisiting} 
                         & 41.9\small±2.0          & 31.3\small±2.1          & 63.7\small±2.0          & 40.6\small±2.2          & 31.8\small±2.5          & 56.2\small±1.8         & \multicolumn{1}{c}{32.7\small±1.9}          \\  
                          
                         & MORE\cite{wang2021more} 
                         & 42.0\small±2.2  	& 43.8\small±1.9  	& 40.3\small±2.0 	& 41.9\small±2.1 	& 40.8\small±2.2 	& 43.1\small±2.4 	& \multicolumn{1}{c}{35.6\small±2.1}  
                         \\  
                         & OHRE\cite{zhang2021open} 
                         & 42.5\small±1.9          & 32.7\small±1.8           & 60.7\small±2.3         & 42.3\small±1.8           & 34.8\small±2.1          & 53.9 \small±2.5        & \multicolumn{1}{c}{33.6\small±1.8 }          \\ 
                         & EIURE\cite{liu2021element} 
                         & 43.1\small±1.8 	& 48.4\small±1.9 	& 38.8\small±1.8 	& 42.7\small±1.6 	& 37.7\small±1.5 	& 49.2\small±1.6  	& \multicolumn{1}{c}{34.5\small±1.4} 
                         \\  
                         
                        & {\modelname} w/o ExemNCE
                         & 40.2\small±1.4         & 37.4\small±1.6          & 43.5\small±1.5          & 39.5\small±1.6          & 34.2\small±1.7          & 46.7\small±1.6          & \multicolumn{1}{c}{32.9\small±1.1}          \\ 
                        & {\modelname} w/o HPC
                         & 41.4\small±1.2         & 38.7\small±1.0          & 44.3\small±0.9          & 41.5\small±1.3          & 37.2\small±1.1          & 47.0\small±0.8          & \multicolumn{1}{c}{34.3\small±0.9}        \\  
                         & {\modelname} w. 10 clusters & 44.3\small±0.5 & 39.9\small±0.6 & 49.8\small±0.5          & 44.9\small±0.4 & 40.0\small±0.5 & 51.2\small±0.4         & \multicolumn{1}{c}{38.3\small±0.6} \\
                         & {\modelname} & \textbf{45.3\small±0.6} & 40.2\small±0.7 & 51.8\small±0.6         & \textbf{45.9\small±0.5} & 40.0\small±0.6 & 53.8\small±0.5         &
                         \multicolumn{1}{c}{\textbf{38.6\small±0.7}} \\
                         \hline \hline
                          
\multirow{9}{*}{TACRED}  
                        & rel-LDA\cite{yao2011structured}          
                        & 35.6\small±2.6 	& 32.9\small±2.5 	& 38.8\small±3.1 	& 38.0\small±3.5 	& 33.7\small±2.6 	& 43.6\small±3.7 & 
                        \multicolumn{1}{c}{21.9\small±2.6}          \\  
                         & March\cite{marcheggiani2016discrete}         
                         & 38.8\small±2.9 	& 35.5\small±2.8 	& 42.7\small±3.2 	& 40.6\small±3.1 	& 36.1\small±2.7 	& 46.5\small±3.2 & 
                         \multicolumn{1}{c}{25.3\small±2.7}          \\  
                         & UIE-PCNN\cite{simon2019unsupervised} 
                         & 41.4\small±2.4 	& 44.0\small±2.7 	& 39.1\small±2.1 	& 41.3\small±2.3 	& 40.6\small±2.2 	& 42.1\small±2.6 & 
                         \multicolumn{1}{c}{30.6\small±2.5}          \\ 
                         & UIE-BERT\cite{simon2019unsupervised}  
                         & 43.1\small±2.0 	& 43.1\small±1.9 	& 43.2\small±2.3 	& 49.4\small±2.1 	& 48.8\small±2.1 	& 50.1\small±2.5 &
                         \multicolumn{1}{c}{32.5\small±2.4}          \\
                         & SelfORE\cite{hu-etal-2020-selfore} 
                         & 47.6\small±1.7 	& 51.6\small±2.0 	& 44.2\small±1.9 	& 52.1\small±2.2 	& 51.3\small±2.0 	& 52.9\small±2.3 & 
                         \multicolumn{1}{c}{36.1\small±2.0}          \\ 
                         & EType\cite{tran-etal-2020-revisiting} 
                         & 49.3\small±1.9 	& 51.9\small±2.1 	& 47.0\small±1.8 	& 53.6\small±2.2 	& 52.5\small±2.1 	& 54.8\small±1.9 & 
                         \multicolumn{1}{c}{35.7\small±2.1}          \\ 
                         
                         & MORE\cite{wang2021more} 
                         & 50.2\small±1.8 	& 56.9\small±2.2 	& 44.9\small±1.8 	& 57.4\small±2.1 	& 56.7\small±1.8 	& 58.1\small±2.3 & 
                         \multicolumn{1}{c}{37.3\small±1.9} 
                         \\  
                         & OHRE\cite{zhang2021open} 
                         & 51.8\small±1.6 	& 55.2\small±2.1 	& 48.7\small±1.7 	& 56.4\small±1.8 	& 55.5\small±1.9 	& 57.3\small±2.1 & 
                         \multicolumn{1}{c}{38.0\small±1.7 }
                         \\ 
                         & EIURE\cite{liu2021element} 
                         & 52.2\small±1.4 	& 57.4\small±1.3 	& 47.8\small±1.5 	& 58.7\small±1.2 	& 57.7\small±1.4 	& 59.7\small±1.7 &
                         \multicolumn{1}{c}{38.6\small±1.1}
                         \\  
                         & {\modelname} w/o ExemNCE
                         & 47.3\small±1.1 	& 51.2\small±1.2 	& 43.9\small±0.9 	& 56.4\small±1.0          & 50.3\small±1.2          & 64.2\small±1.4           & \multicolumn{1}{c}{36.9\small±1.0}          \\ 
                        & {\modelname} w/o HPC
                          & 48.4\small±0.9          & 50.3\small±0.8          & 46.7\small±1.2          & 58.1\small±1.1          & 51.8\small±1.4          & 66.2\small±1.5          & \multicolumn{1}{c}{37.8\small±0.8}          \\ 
                         & {\modelname} w. 10 clusters
                         & 55.8\small±0.4 	& 57.8\small±0.3 	& 54.0\small±0.5 	& 59.7\small±0.6 	& 57.6\small±0.5 	& 61.9\small±0.6 & \multicolumn{1}{c}{40.5\small±0.4} \\
                        & {\modelname} & \textbf{56.7\small±0.4} & 58.4\small±0.5 & 55.0\small±0.3        & \textbf{61.3\small±0.5} & 59.5\small±0.6 & 63.1\small±0.4        & \multicolumn{1}{c}{\textbf{42.2\small±0.5}} \\ \thickhline
\end{tabular}
}

\caption{Quantitative performance evaluation on two datasets.}\label{tab:data}
\vspace{-0.1in}
\end{table*}

\subsection{Datasets}
Following previous work \cite{simon2019unsupervised,hu-etal-2020-selfore,tran-etal-2020-revisiting}, we employ NYT+FB to train and evaluate our model. The NYT+FB dataset is generated via distant supervision, aligning sentences from the New York Times corpus \cite{sandhaus2008new} with Freebase \cite{bollacker2008freebase} triplets. We follow the setting in \citet{hu-etal-2020-selfore,tran-etal-2020-revisiting} and filter out sentences with non-binary relations. We get 41,685 labeled sentences containing 262 target relations (including \textit{no\_relation}) from 1.86 million sentences.

There are two more further concerns when we use the NYT+FB dataset, which is also raised by \citet{tran-etal-2020-revisiting}. Firstly, the development and test sets contain lots of wrong/noisy labeled instances, where we found that more than 40 out of 100 randomly selected sentences were given the wrong relations. Secondly, the development and test sets are part of the training set. Even under the setting of unsupervised relation extraction, this is still not conducive to reflect the performance of models on unseen data. Therefore, we follow \citet{tran-etal-2020-revisiting} and additionally evaluate all models on the test set of TACRED \cite{zhang2017position}, a large-scale crowd-sourced relation extraction dataset with 42 relation types (including \textit{no\_relation}) and 18,659 relation mentions in the test set.

\subsection{Baselines}
We use standard unsupervised evaluation metrics for comparisons with other eight baseline algorithms:
1) \textbf{rel-LDA} \cite{yao2011structured},  generative model that considers the unsupervised  relation  extraction  as  a  topic  model. We choose the full rel-LDA with a total number of 8 features for comparison. 2) \textbf{MARCH}\cite{marcheggiani2016discrete} proposed a discrete-state variational autoencoder (VAE) to tackle URE. 3) \textbf{UIE} \cite{simon2019unsupervised} trains a discriminative RE model on unlabeled instances by forcing the model to predict each relation with confidence and encourages the number of each relation to be predicted on average, where two base models (UIE-PCNN and UIE-BERT) are considered. 4) \textbf{SelfORE} \cite{hu-etal-2020-selfore} is a self-supervised framework that clusters self-supervised signals generated by BERT adaptively and bootstraps these signals iteratively by relation classification. 5) \textbf{EType} \cite{tran-etal-2020-revisiting} uses one-hot vector of the entity type pair to ascertain the important features in URE. 6) \textbf{MORE} \cite{wang2021more} utilizes deep metric learning to obtain rich supervision signals from labeled data and drive the neural model to learn semantic relational representation directly. 7) \textbf{OHRE} \cite{zhang2021open} proposed a dynamic hierarchical triplet objective and hierarchical curriculum training paradigm for open relation extraction. 8) \textbf{EIURE} \cite{liu2021element} is the state-of-the-art method that intervenes on the context and entities respectively to obtain the underlying causal effects of them. Since most of the baseline methods do not exactly match the dataset and experimental setup of our method, the baselines are reproduced and adjusted to the same setting to ensure a fair comparison.

\subsection{Results}

Since most baseline methods adopted the setting by clustering all samples into 10 relation classes \cite{simon2019unsupervised,hu-etal-2020-selfore,tran-etal-2020-revisiting,liu2021element}, we adjust the $p_{\bot}$ in Algorithm \ref{alg:pc_algorithm} to get the same results  for fair comparison, and name this setting {\modelname} w. 10 clusters. Although 10 relation classes are lower than the number of true relation types in the dataset, it still reveals important insights about models' ability to tackle skewed distribution. 

Table \ref{tab:data} demonstrates the average performance and standard deviation of the three runs of our model in comparison with the baselines on NYT+FB and TACRED. We can observe that EIURE achieves the best performance among all the baselines, which is considered as the previous state-of-the-art method. The proposed {\modelname} outperforms all baseline models consistently on B$^{3}$ F1, V-measure F1, and ARI. {\modelname} on average achieves 3.4\% higher in B$^{3}$ F1, 2.9\% higher in V-measure F1, and 3.9\% higher in ARI on two datasets when comparing with EIURE. The standard deviation of {\modelname} is particularly lower than other baseline methods, which validates its robustness.
Furthermore, the performance of {\modelname} on TACRED exceeds all the baseline methods by at least 2.1\%. These performance gains are likely from both 1) higher-quality manually labeled samples in TACRED and 2) an improved discriminative power of {\modelname} considering the variation and semantic shift from NYT+FB to TACRED.

\noindent\textbf{Effectiveness of HPC.}\quad 
HPC considers all data points and uses their mutual similarity to find the most suitable points as exemplars for each cluster, these exemplars could update the instances in their own clusters and transfer the relational features from high-level relations to base-level through the cross hierarchy attention. From Table \ref{tab:data}, {\modelname} w/o HPC, which uses $k$-means instead of the proposed hierarchical clustering, gives 4.7\% less performance in average over all metrics when comparing with {\modelname}. 

\begin{figure}[bt!]
\subfigure[Effect of CHA]{
\begin{minipage}[t]{0.48\linewidth}
\centering
\includegraphics[width=1\linewidth,height=0.7\textwidth]{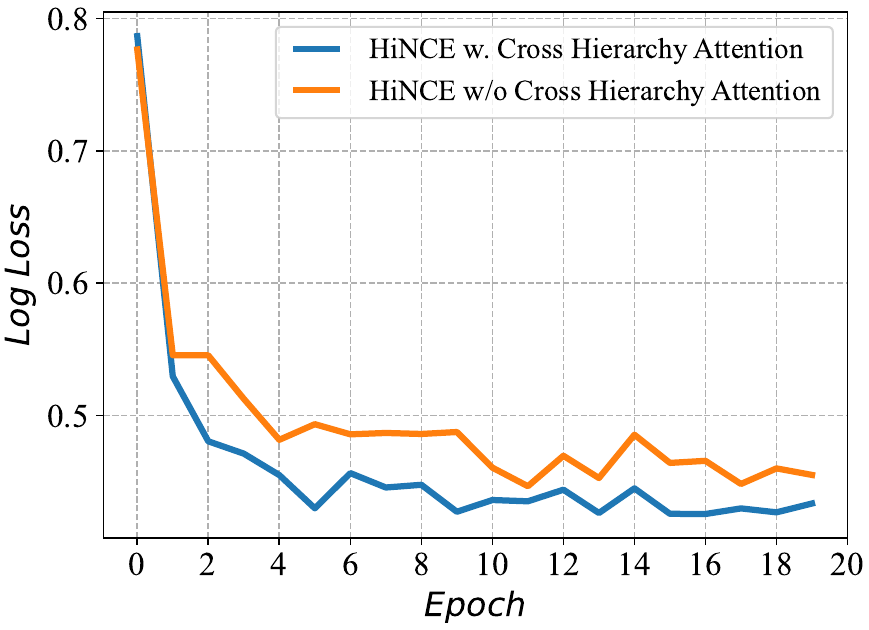}
\end{minipage}%
\label{fig:log_loss}
}%
\subfigure[Effect of HiNCE]{
\begin{minipage}[t]{0.48\linewidth}
\centering
\includegraphics[width=1\linewidth,height=0.7\textwidth]{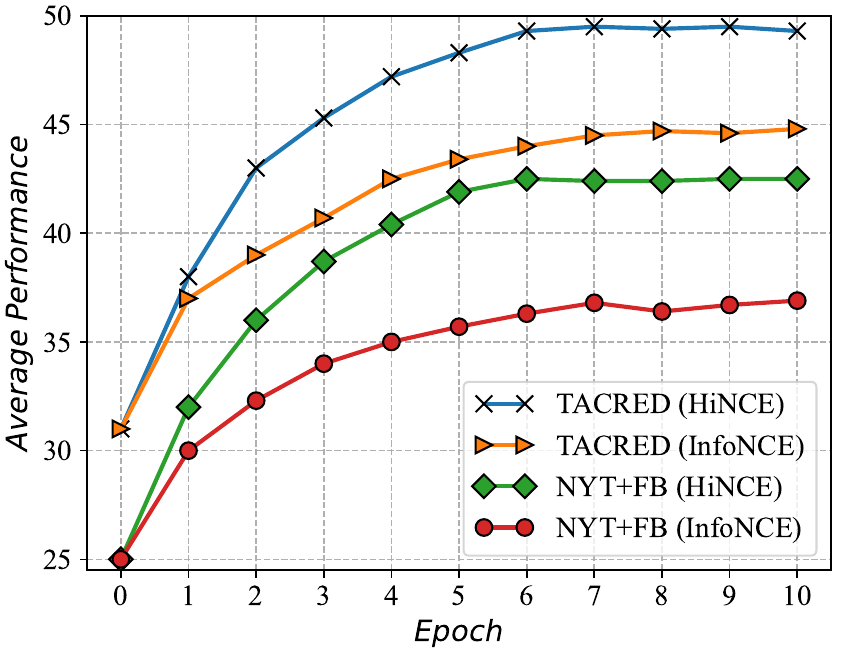}
\end{minipage}%
\label{fig:epoch performance}
}%

\vspace{-0.1in}
\caption{Effect of Cross Hierarchy Attention on NLL loss on NYT+FB dataset (left) and HiNCE on average performance of two datasets (right) while training.}\label{fig:performnce_loss}
\end{figure}

\begin{figure}[bt!]
    \centering
    \includegraphics[width=0.95\linewidth]{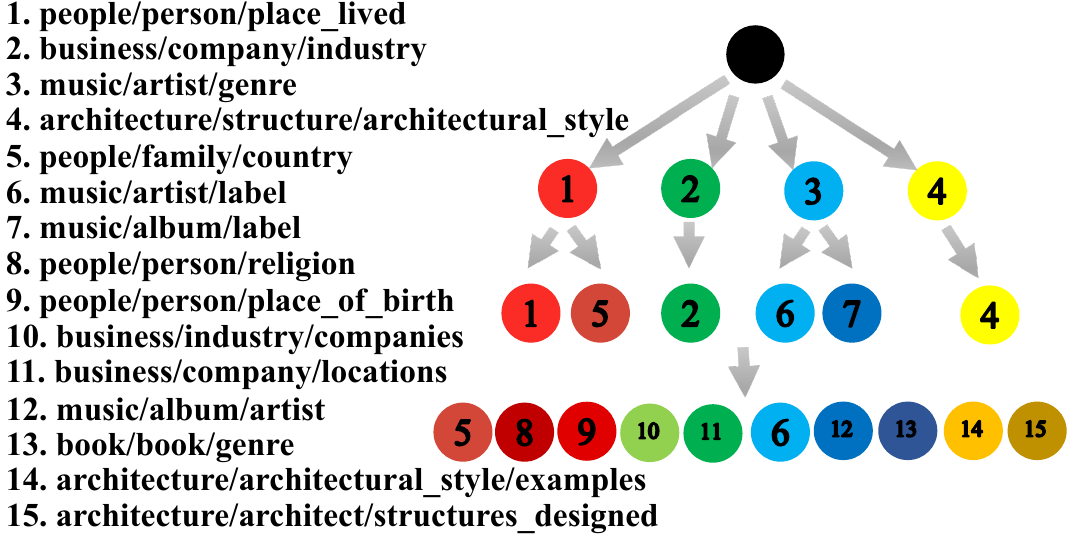}
    \caption{Relation hierarchy derived from the feature space on the NYT+FB dataset.}
    \label{fig:hierarchical relation}
\vspace{-0.05in}
\end{figure}

\noindent\textbf{Effectiveness of Cross Hierarchy Attention.}\quad 
In order to explore how CHA helps data points to obtain the semantics of exemplars as training signals in HPC, Figure \ref{fig:log_loss} illustrates the log loss values of HiNCE during the training epochs. Based on the loss curve, using Cross Hierarchy Attention leads to consistently lowered loss value, which implies that it provides high-quality signals to help train a better relational clustering model. 

Considering that our exemplars correspond to specific data points and relations, we further show the hierarchical relations the model derived from the dataset. From Figure \ref{fig:hierarchical relation}, we can observe a three-layer exemplars structure the model derives from the NYT+FB dataset without any prior knowledge.
The high-level relations and base-level relations belonging to an original cluster convey similar relation categories, which demonstrates the rationality of exemplars in relational feature clustering. As the number of exemplars between different layers increases, some exemplars are adaptively replaced with more fine-grained ones in the base-level layer. 

Note that the approach in this paper works best only when the relational structure in the dataset is hierarchical. Other cases, such as graph structures or binary structures, are untested and may not perform optimally.

\noindent\textbf{Effectiveness of HiNCE.}\quad 
The main purpose of HiNCE is to leverage exemplar-wise contrastive learning in addition to instance-wise. HiNCE avoids the pitfall where many instance-wise negative pairs share similar semantics but are undesirably pushed apart in the feature space. We first conduct an ablation study to demonstrate the effectiveness of this module. From Table \ref{tab:data}, {\modelname} w/o HiNCE gives us 6.3\% less performance averaged over all metrics. Then we report the average performance of B$^{3}$ F1, V-measure F1, and ARI on the two datasets changing with epochs, which reflects the quality and purity of the clusters generated by {\modelname}. From Figure \ref{fig:epoch performance}, compared to InfoNCE alone, training on HiNCE can improve the performance as training epochs increase, indicating that better representations are obtained to form more semantically meaningful clusters.
\begin{figure}[bt!]
\subfigure[{\modelname} (higher-level relations)]{
\begin{minipage}[t]{0.48\linewidth}
\centering
\includegraphics[width=1\linewidth]{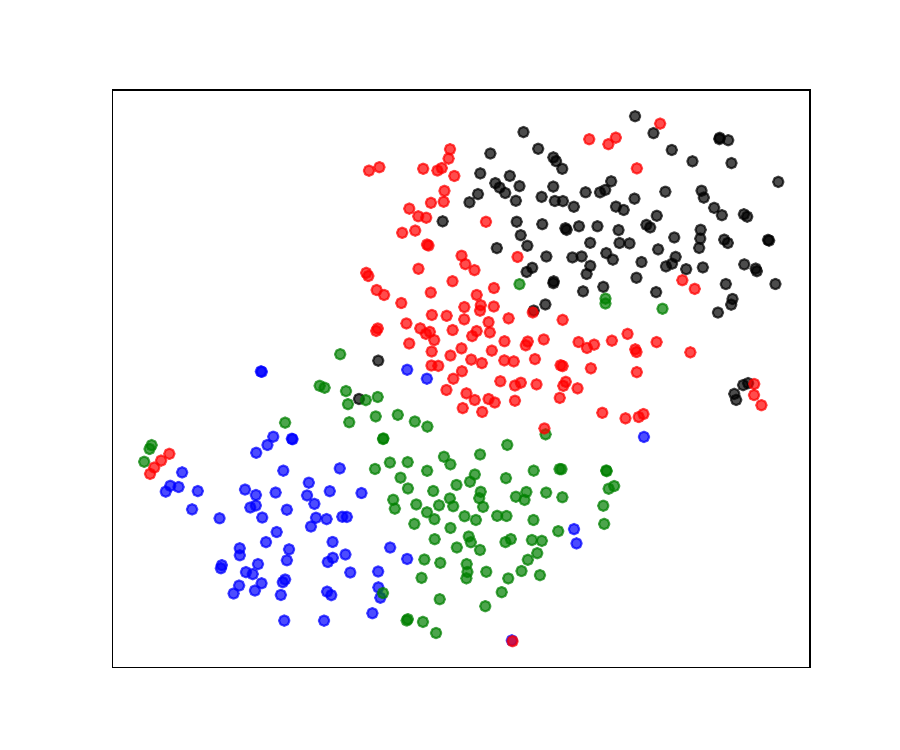}
\end{minipage}%
}%
\subfigure[{\modelname}]{
\begin{minipage}[t]{0.48\linewidth}
\centering
\includegraphics[width=1\linewidth]{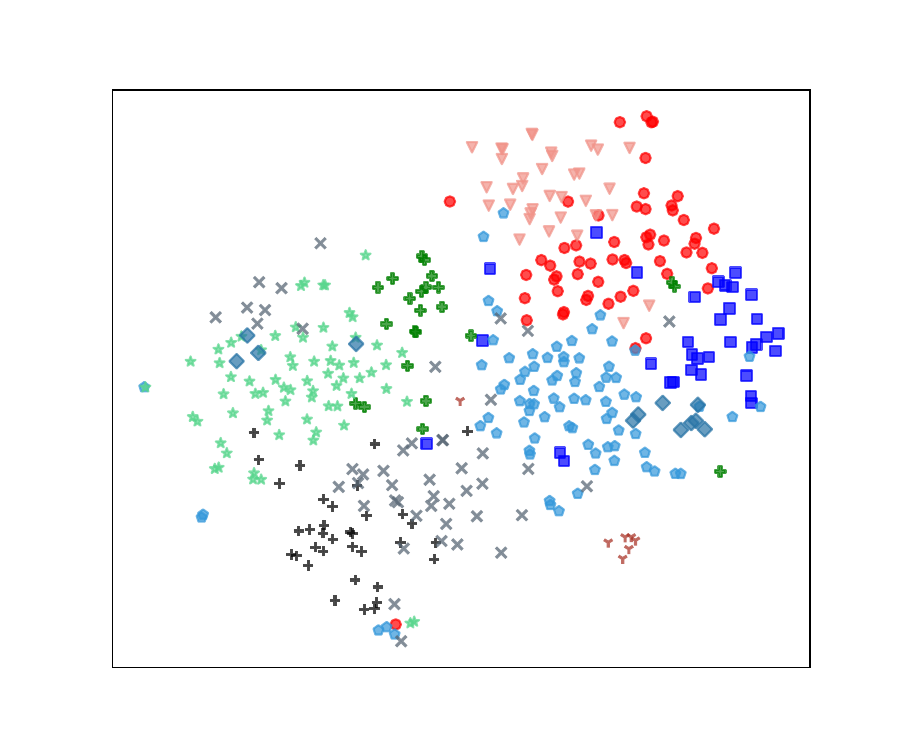}
\end{minipage}%
}%

\subfigure[{\modelname} w/o ExemNCE]{
\begin{minipage}[t]{0.48\linewidth}
\centering
\includegraphics[width=1\linewidth]{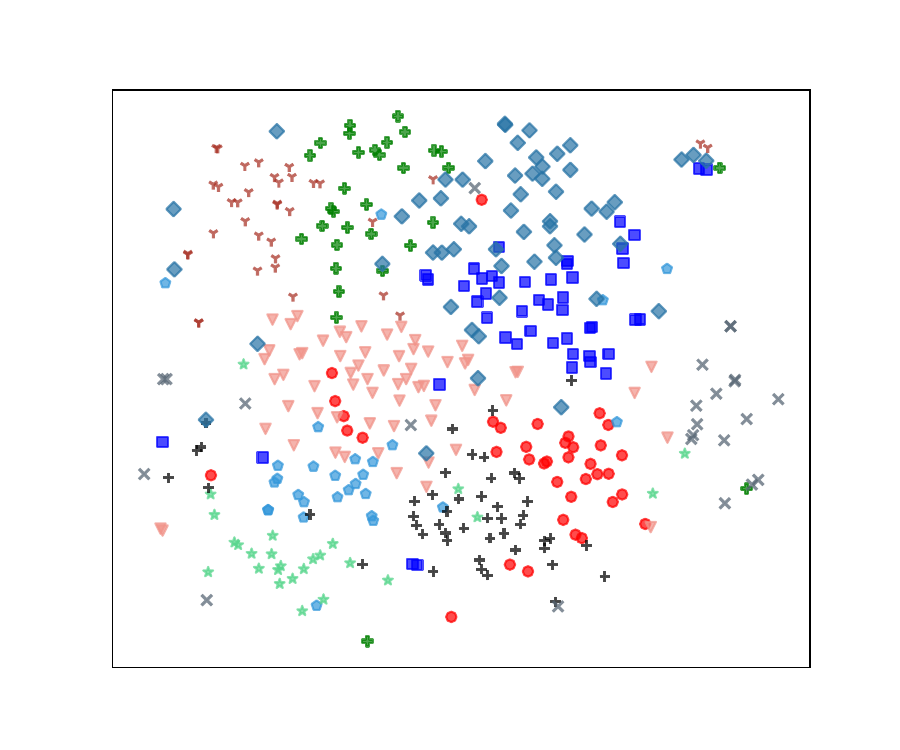}
\end{minipage}%
}%
\subfigure[{\modelname} w/o HPC]{
\begin{minipage}[t]{0.48\linewidth}
\centering
\includegraphics[width=1\linewidth]{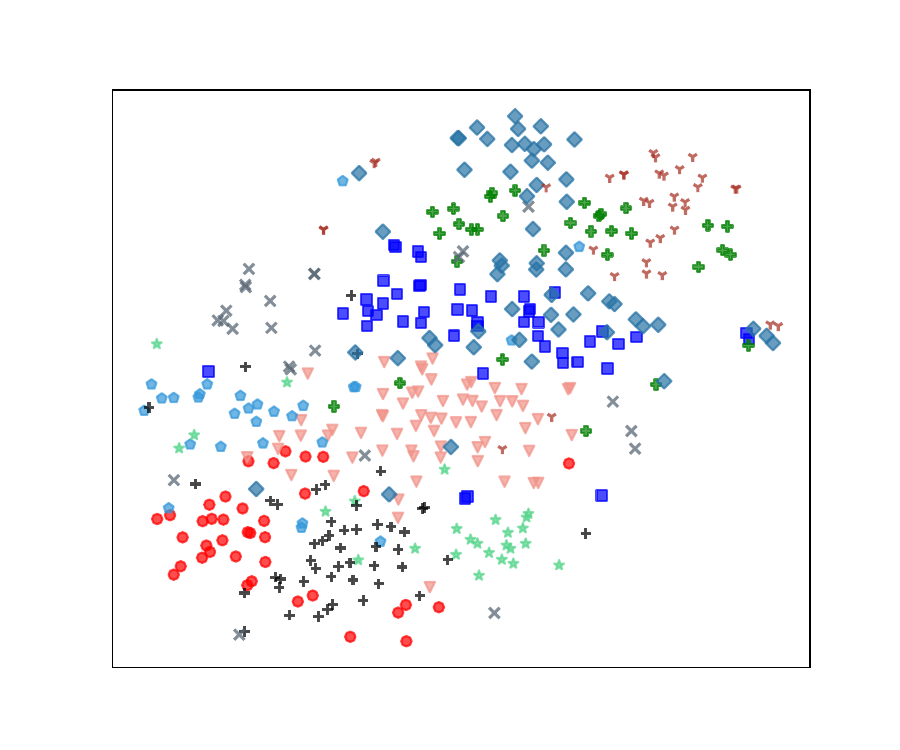}
\end{minipage}%
}%
\vspace{-0.1in}
\caption{Visualizing contextualized entity-level features after t-SNE dimension reduction on TACRED dataset.}\label{fig:tsne}
\vspace{-0.1in}
\end{figure}
\noindent\textbf{Visualize Hierarchical Contextualized Features.}
To further intuitively show how tree-structured hierarchical exemplars help learn better contextualized relational features on entity pairs for URE, we visualize the contextual representation space $\mathbb{R}^{\left(2+P\right)\cdot{b_{R}}}$ after dimension reduction using t-SNE \cite{maaten2008visualizing}. We randomly choose 400 relations from TACRED dataset and the visualization results are shown in Figure \ref{fig:tsne}.

From Figure \ref{fig:tsne} (a), we can see that {\modelname} can give proper clustering results to the higher-level relational features generated by propagation clustering, where features are colored according to their clustering labels. In order to explore how our modules utilize high-level relation features to guide the clustering of base-level relations, we preserve the color series of the corresponding high-level clustering relation labels, while base-level clustering relation labels with different shapes to get Figure \ref{fig:tsne} (b) (c) (d). {\modelname} in (b) learns denser clusters and discriminitaive features. However, {\modelname} without ExemNCE in (c) is difficult to obtain the semantics of the sentences without exemplar-wise information, which makes the clustering results loose and error-prone. When Hierarchical Propagation Clustering is not applied as (d), $k$-means is adopted to perform clustering on the high-level relational features, which could not use exemplars to update relational features or mutual similarity between feature points. 
On that occasion, {\modelname} w/o HPC gives the results where the points between clusters are more likely to be mixed. The outcomes revealed above prove the effectiveness of {\modelname} to obtain the semantics of sentences while distinguishing between similar and dissimilar sentences.

\noindent\textbf{Hyperparameter Analysis.}\quad 
We have explicitly introduced two hyperparameters $P$ in the encoder and $L$ in the HPC algorithm. We first study the number of Random Span words $P$ which affects the fixed-length of relation representation in Eq. \ref{h} by changing $P$ from 1 to 4 and report the average performance of B$^{3}$ F1, V-measure F1, and ARI on NYT+FB and TACRED. From Table \ref{hyper_exp}, the fluctuation results indicate that both information deficiency and redundancy of relation representations will affect the model's performance. Using short span words will introduce less-information relational features so that is hard to transfer representations from a large scale of sentences, while long span words will cause high computational complexity and lead to information redundancy. 

Then, we study the level of $L$ hierarchical layers as well as the way of building preference sequence to form them, so as to discover the most suitable tree-structured hierarchical relations for the data distribution. We change $L$ from 2 to 5 with fixed top preference $p_{\top}$ and bottom preference $p{_{\bot}}$ to get the effect of $L$ and report the average performance in Table \ref{hyper_exp}. The fluctuation here implies that fewer layers fail to transfer more information while more layers may cause exemplar-level information conflicts between different coarse-grained layers. \cite{moiane2018evaluation} has shown that the minimum and median value of similarity matrix are best preferences for propagation clustering, so we manually adjust the preference sequence between them multiple times with $L=3$ and get the average results as \texttt{3+M} to compare with the automatically generated ones by Line~\ref{line:p_top_and_bot}-\ref{line:generate_prefs} in HPC. The results show that the bottom layer is not so sensitive to the preference sequence as long as it is reasonable, which proves the practicability and effectiveness of the equation in Line \ref{line:generate_prefs}.
\begin{table}[]
\centering
\begin{minipage}{1.0\linewidth}
\centering
\resizebox{.9\linewidth}{!}{
\begin{tabular}{cccccc}
    \hline
    Dataset\ / \ $P$    & 1    & 2    & 3    & 4   & 5    \\ \hline
 NYT+FB & 40.9 & \textbf{42.5} & 41.3 & 40.6  & 39.2\\
 TACRED & 51.2 & \textbf{52.4} & 51.4 & 50.6  & 49.8\\\hline
\end{tabular}
}
\end{minipage}
\begin{minipage}{1.0\linewidth}
\centering
\resizebox{.9\linewidth}{!}{
\begin{tabular}{cccccc}
    Dataset \ / \ $L$    & 2    & 3    & 4 & 5   & 3+M      \\ \hline
 NYT+FB & 38.6 & \textbf{42.5} & 40.9 & 39.2   & \textbf{42.5}\\
 TACRED & 48.8 & \textbf{52.4} & 50.4 & 49.6   & 52.1\\\hline
\end{tabular}
}
\end{minipage}
\caption{Average performance with different number of $P$ and $L$ on NYT+FB and TACRED.}\label{hyper_exp}
\vspace{-0.15in}
\end{table}

\section{Related Work}

Unsupervised relation extraction has received attention recently \cite{simon2019unsupervised,tran-etal-2020-revisiting,hu-etal-2020-selfore}, due to the ability to discover relational knowledge without access to annotations or external resources. Unsupervised models either 1) cluster the relation features extracted from the sentence, or 2) make more assumptions as learning signals to discover better relational representations.

Among clustering models, an important milestone is the self-supervised learning approach \cite{wiles2018self,caron2018deep,hu-etal-2020-selfore}, assuming the cluster assignments as pseudo-labels and a classification objective is optimized. However, these works heavily rely on a frequently re-initialized linear classification layer which interferes with representation learning. \citet{zhan2020online} proposes Online Deep Clustering that performs clustering and network update simultaneously rather than alternatingly to tackle this concern, however, the noisy pseudo labels still affect feature clustering when updating the network \cite{hu2021semi, li2022graph, lin2022inferring}.

Inspired by the success of contrastive learning in computer vision tasks \cite{he2020momentum,li2020prototypical,caron2020unsupervised}, instance-wise contrastive learning in information extraction tasks \cite{peng2020learning, li2022pair}, and large pre-trained language models that show great potential to encode meaningful semantics for various downstream tasks \cite{devlin2019bert,soares2019matching,hu2021gradient}, we proposed a hierarchical exemplar contrastive learning schema for unsupervised relation extraction. It has the advantages of supervised learning to capture high-level semantics in the relational features instead of exploiting base-level sentence differences to strengthen discriminative power and also keeps the advantage of unsupervised learning to handle the cases where the number of relations is unknown in advance.
\section{Conclusion}
In this paper, we propose a contrastive learning framework model {\modelname} for unsupervised relation extraction. Different from conventional self-supervised models which either endure gradual drift or perform instance-wise contrastive learning without considering hierarchical relation structure, our model leverages HPC to obtain hierarchical exemplars from relational feature space and further utilizes exemplars to hierarchically update relational features of sentences and is optimized by performing both instance and exemplar-wise contrastive learning through HiNCE and propagation clustering iteratively. Experiments on two public datasets show the effectiveness of {\modelname} over competitive baselines.

\section{Acknowledgement}
We thank the reviewers for their valuable comments. The work was supported by the National Key Research and Development Program of China (No. 2019YFB1704003), the National Nature Science Foundation of China (No. 62021002 and No. 71690231), NSF under grants III-1763325, III-1909323, III-2106758, SaTC-1930941, Tsinghua BNRist and Beijing Key Laboratory of Industrial Bigdata System and Application.

\bibliography{acl2021}
\bibliographystyle{acl_natbib}

\newpage
\appendix
\section{Implementation Details}\label{impl_details}
In the encoder phase, we set the number $P$ of randomly selected words in the $[Span]$ to 2, the reason of which is illustrated in parameter analysis. Therefore the output entity-level features $\mathbf{h}_{i}$ and $\mathbf{h}_{i}^{\prime}$ possess the dimension of ${4\cdot{b_{R}}}$, where $b_R=768$. We use the pretrained \texttt{BERT-Base-Cased} model to initialize both the Momentum Encoder and Propulsion Encoder respectively, and use AdamW \cite{loshchilov2017decoupled} to optimize the loss. The encoder is trained for 20 epochs with $1e{-5}$ learning rate. 
In the HPC phase, we set the numbers of layers $L$ to 3 after parameter analysis and the maximum iterations at Line \ref{line:iterations_start} to 400 to make sure the algorithm terminates in time and make the converge condition as $E^l$ not change for 10 iterations. We set temperature parameter $\tau=0.02$ and momentum parameter $m=0.999$ following \cite{he2020momentum} and adjust the number of negative samples $J$ to $512$ to accommodate smaller batches.

\section{Evaluation metrics}\label{evaluation_metrics}
 We follow previous works and use ${\text{B}^{3}}$ \cite{bagga1998entity}, V-measures \cite{rosenberg2007v} and Adjusted Rand Index (ARI) \cite{hubert1985comparing} as our end metrics.
${\text{B}^{3}}$ uses precision and recall to measure the correct rate of assigning data points to its cluster or clustering all points into a single class. 
We use V-measures \cite{rosenberg2007v} to calculate homogeneity and completeness, which is analogous to ${\text{B}^{3}}$ precision and recall.
These two metrics penalize small impurities in a relatively ``pure'' cluster more harshly than in less pure ones. We also report the F1 value, which is the harmonic mean of Hom. and Comp.
Adjusted Rand Index (ARI) \cite{hubert1985comparing} measures the similarity of predicted and golden data distributions. The range of ARI is [-1,1]. The larger the value, the more consistent the clustering result is with the real situation.

\end{document}